\documentclass[10pt,twocolumn,letterpaper]{article}
\usepackage[utf8]{inputenc}
 
\usepackage{cvpr}
\usepackage{times}
\usepackage{epsfig}
\usepackage{graphicx}
\usepackage{amsmath}
\usepackage{amssymb}
\usepackage{capt-of}
\usepackage{subcaption}
\usepackage{mdframed}

\usepackage[pagebackref=true,breaklinks=true,letterpaper=true,colorlinks,bookmarks=false]{hyperref}

\cvprfinalcopy 

\def\cvprPaperID{****} 

\begin{document}

\title{Who is Mistaken?}

\author{Benjamin Eysenbach\\
MIT\\
{\tt\small bce@mit.edu}
\and
Carl Vondrick\\
MIT\\
{\tt\small vondrick@mit.edu}
\and
Antonio Torralba\\
MIT\\
{\tt\small torralba@csail.mit.edu}
}


\maketitle
\thispagestyle{empty}

\enlargethispage{-5.5cm}
\noindent\begin{picture}(0,0)
\put(0,-430){\begin{minipage}{\textwidth}
\centering
\includegraphics[width=\linewidth]{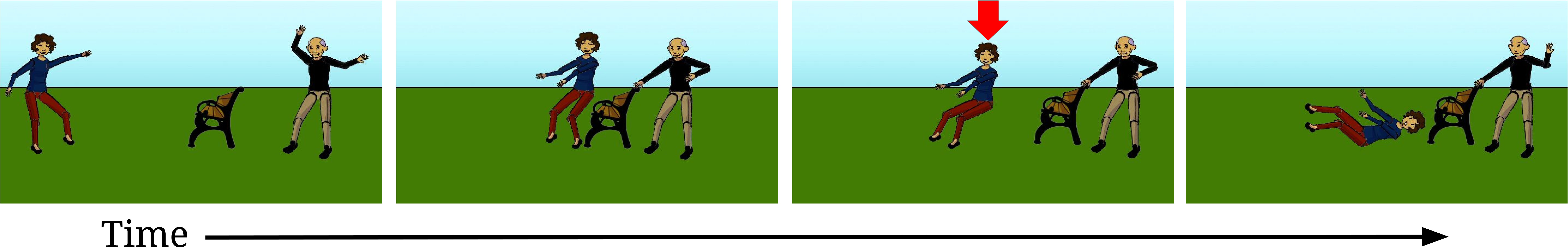}
\captionof{figure}{\textbf{Can you determine who believes something incorrectly in this scene?} In this paper, we study how to recognize when a person in a scene is mistaken. Above, the woman is mistaken about the chair being pulled away from her in the third frame, causing her to fall down. The \textbf{\textcolor{red}{red arrow}} indicates false belief. We introduce a new dataset of abstract scenes to study when people have false beliefs. We propose approaches to learn to recognize \textbf{who} is mistaken and \textbf{when} they are mistaken. 
}
\label{fig:teaser}
\end{minipage}}
\end{picture}%

\begin{abstract}
\vspace{-1em}
Recognizing when people have false beliefs is crucial for understanding their actions. We introduce the novel problem of identifying when people in abstract scenes have incorrect beliefs. We present a dataset of scenes, each visually depicting an 8-frame story in which a character has a mistaken belief. We then create a representation of characters' beliefs for two tasks in human action understanding: predicting who is mistaken, and when they are mistaken. Experiments suggest that our method for identifying mistaken characters performs better on these tasks than simple baselines. Diagnostics on our model suggest it learns important cues for recognizing mistaken beliefs, such as gaze. We believe models of people's beliefs will have many applications in action understanding, robotics, and healthcare.
\vspace{-1em}
\end{abstract}

\section{Introduction}

In Figure \ref{fig:teaser}, one person has a mistaken belief about their environment. Can you figure out who is mistaken? You likely can tell the woman is about to sit down because she incorrectly believes the chair is there. Although you can see the complete scene, the character inside the scene has an imperfect view of the world, causing an incorrect belief. 

\begin{figure*}
\centering
\includegraphics[width=\textwidth]{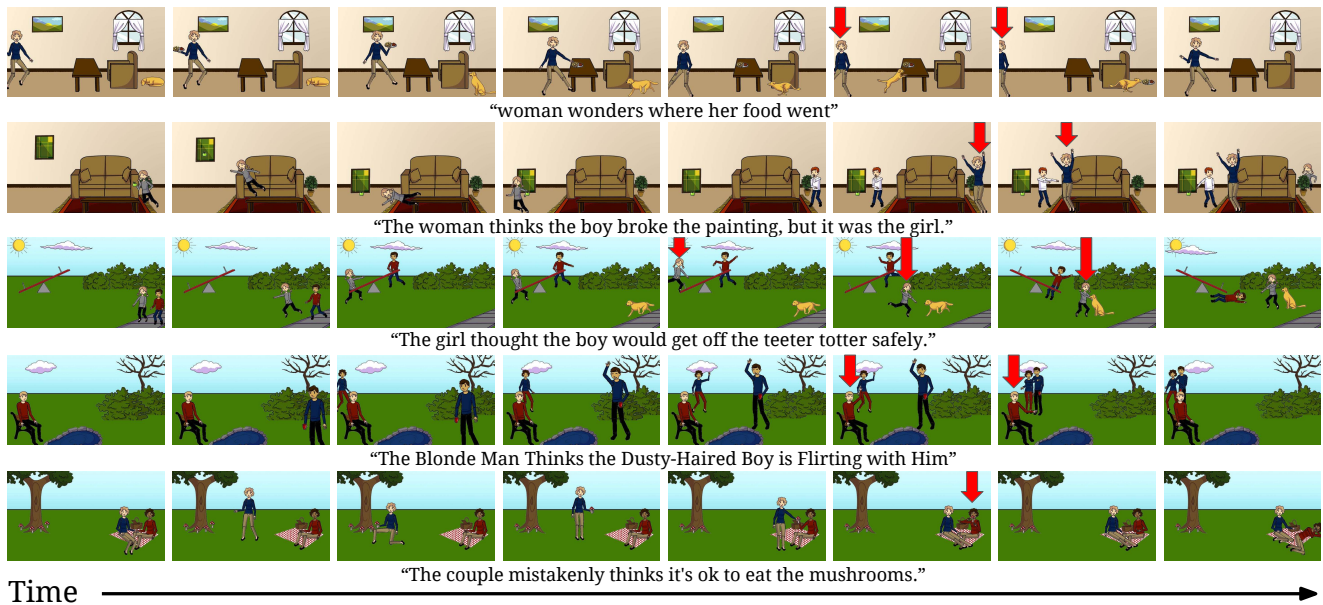}
\vspace{-1.5em}
\caption{\textbf{Visual Beliefs Dataset:} We introduce a new dataset of abstract scenes to study visual beliefs. We show five example scenes from our dataset. The \textbf{\textcolor{red}{red arrows}} indicate that a person has a false belief in that frame. Each scene (row) contains eight images, depicting a visual story when read left to right. The caption below each scene was collected during annotation for visualization purposes only.}
\vspace{-1.5em}
\label{fig:grid}
\end{figure*}

The ability to recognize when people have incorrect beliefs will enable several key applications in computer vision, such as in action understanding, robotics, and healthcare. For example, understanding beliefs of human drivers could improve the safety of autonomous vehicles \cite{sadigh2016information}. Robots that understand human beliefs may have more fluid interactions with humans \cite{koppula2013learning}. Understanding beliefs may provide clues for anticipating human actions \cite{kitani2012activity,vondrickanticipating} and generate better visual humor \cite{humor}. How do we give machines the capability to understand what a person believes? 

\enlargethispage{-5.5cm}
In this paper, 
we introduce the novel problem of recognizing incorrect beliefs in short visual stories. We propose two new tasks aimed at understanding which people have false beliefs. Given a visual story, we aim to recognize \textbf{who} is mistaken and \textbf{when} they are mistaken. For example, in Figure \ref{fig:teaser}, the woman is mistaken in the third frame.

To study this problem, we present a dataset of abstract scenes \cite{zitnick2013bringing} that depict visual stories of people in various types of everyday situations. In each story, one or more people have mistaken beliefs, and we seek to recognize these people. Abstract scenes are ideal for studying this problem because we can economically create large datasets that focus on the human activities, such as ones influenced by people's beliefs. Moreover, while abstract scenes are synthetic, the data models  behavior on a high-level and can be applied to natural images with domain adaptation. The scenarios in our dataset are diverse and characters are mistaken for many reasons, such as occlusion or unexpected actions.


We investigate models for learning to recognize mistaken characters in short sequences.
Our model uses person-centric representations of scenes and combines information across several timesteps to better recognize mistaken characters.
Experiments show that our model learns to mistaken people beliefs better than baselines, suggesting that it is possible to make progress on inferring people's beliefs.
Although we only train our model to predict mistaken beliefs, experiments suggest that it internally learns important cues for beliefs, such as human gaze or time's arrow.

The first contribution of this paper is introducing two new computer vision tasks for recognizing beliefs in images. 
The second contribution is a new dataset for training and evaluating models for recognizing beliefs. The third contribution is a  model for starting to tackle these belief tasks. Code, data, and models will be available at \mbox{\small{\url{http://people.csail.mit.edu/bce/mistaken/}}}.

\section{Related Work}
\label{sec:related-work}

\textbf{Beliefs and Intentions:}
Our paper builds off several works that study beliefs of people.
Shepherd \cite{psych-gaze} studies humans' \emph{theory of mind}, their reasoning about beliefs of others. He notes that gaze-following is important for this reasoning and failing to solve this problem may indicate a disability.  Scassellati \cite{scassellati2002theory} studies theory of mind in human-robot interaction.  Xie et al.\ \cite{xie2013inferring} explore people's intentions in real-world surveillance footage. Baker et al.\ \cite{bayesian-tom} propose a Bayesian model for learning beliefs based on a POMDP.  Zhao et al.\ \cite{tom-hri} propose using probabilistic programming to infer the beliefs and desires of people in RGBD videos.
We focus on learning the beliefs of characters directly from visual scenes.


\begin{figure*}
\centering
    \begin{subfigure}[t]{0.28\textwidth}
        \centering
        \captionsetup{width=.75\linewidth}
        \includegraphics[width=0.85\textwidth]{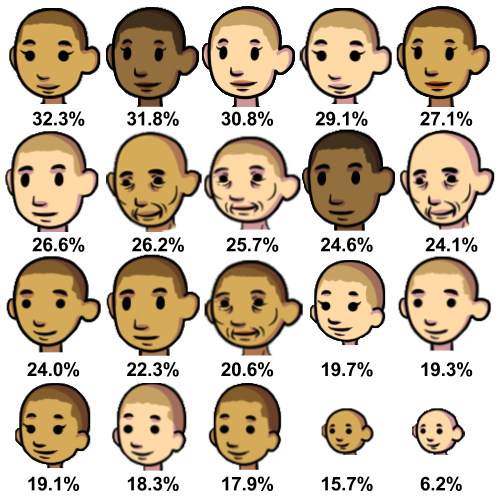}
        \caption{\textbf{Character ID:} For the 20 characters in our dataset, we show the probability they are mistaken in frames where each is present.}
    \end{subfigure}%
    ~ 
    \begin{subfigure}[t]{0.11\textwidth}
        \centering
        \captionsetup{width=1.7\linewidth,oneside,margin={-1.5em,-1.5em}}
        \includegraphics[width=0.85\textwidth]{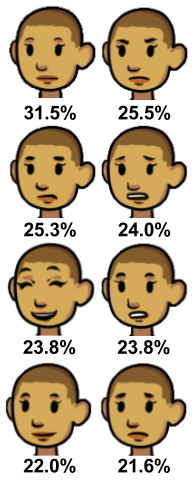}
        \caption{\textbf{Facial expressions:} We show the probability a character is mistaken given their facial expression.}
    \end{subfigure}
    ~
    \begin{subfigure}[t]{0.23\textwidth}
        \centering
        \captionsetup{width=.6\linewidth,oneside,margin={3.2em,0.4em}}
        \includegraphics[width=0.9\textwidth]{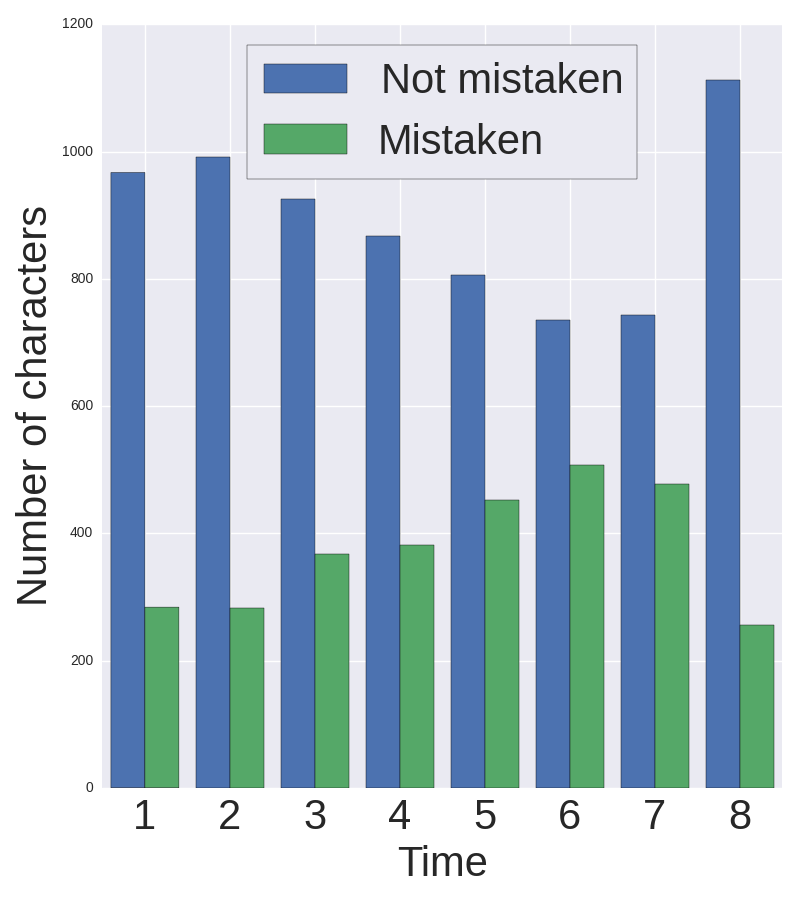}
        \caption{\textbf{Time:} People tend to be mistaken towards the end of the scene.}
    \end{subfigure}
    ~
    \begin{subfigure}[t]{0.33\textwidth}
        \centering
        \captionsetup{width=1.0\linewidth}
        \includegraphics[width=0.9\textwidth]{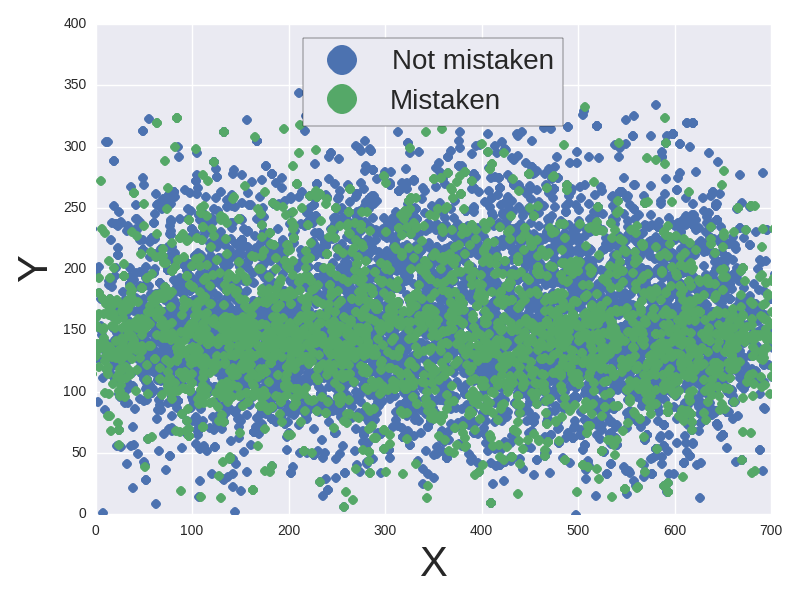}
        \caption{\textbf{Location:}  We show the $(x, y)$ location of every character in every frame. The distribution for mistaken characters and not-mistaken characters appears similar.}
    \end{subfigure}
    \vspace{-0.5em}
    \caption{\textbf{Dataset Statistics:} We summarize biases of mistaken characters. Our method performs better than baselines that exploit these biases (see Table~\ref{table:main_experiment}). \label{fig:data}}
    \vspace{-1.5em}
\end{figure*}

\textbf{Common Sense:}
Our work complements efforts to learn common sense.
Yatskar et al.\ \cite{yatskarstating} extract common sense from object detection corpora, while Chen et al.\ \cite{chen2013neil} learn visual common sense by browsing the Internet.  Vedantam et al.\ \cite{abstract-commonsense} use abstract images to learn how people, animals and objects are likely to interact. Recent work \cite{block-towers, galileo, pinto2016curious} has learned physical common sense given videos of colliding objects. Finally, Alahi et al.\ \cite{alahisocial} explore understanding social interactions in crowded spaces, and Prabhakar et al.\ \cite{prabhakar2010temporal} study causality in unconstrained video to understand social games. In this work, we study the subset of common sense related to visual beliefs.

\textbf{Activity Understanding:} Our work is  related to activity understanding in vision \cite{caba2015activitynet,wang2011action,chao2015hico,pirsiavash2012detecting,fathi2012learning}. Systems for understanding human actions typically leverage a variety of cues, such as context, pose, or gaze \cite{recasens2015they}. Our work complements action understanding in two ways. First, we study visual beliefs, which may be a useful signal for better understanding people's activities. Second, recognizing visual beliefs often requires an understanding of people's actions. 

\textbf{Abstract Images:}
We take advantage of abstract images pioneered by Zitnick et al.\ \cite{zitnick2013bringing}, which have received wide interest in computer vision for studying high-level vision tasks.
Chandrasekaran et al.\ \cite{humor} use abstract images to detect visual humor. Zhang et al.\ \cite{zhang2015yin} explore binary question-answering in abstract scenes, and Fouhey et al.\ \cite{fouhey2014predicting} learn to predict object dynamics in clip art. While these approaches reason about image-level features and semantics, our approach looks at character-level features. Importantly, two characters in the same scene can have different beliefs about the world, so each character should have a different character-level feature. Additionally, we extend this previous work to multi-frame scenes depicting visual stories.

\textbf{Transfer:}
After we learn to recognize mistaken characters in abstract scenes, one could use domain adaptation \cite{fouhey2014predicting, castrejon2016learning} to apply our approach to natural images. However, this is orthogonal to the goal of this paper. Additionally, Ganin et al.\ \cite{ganin2014unsupervised} and Tzeng et al.\ \cite{tzeng2015simultaneous} show how to perform unsupervised domain adaptation, which is relevant to our setting because annotating natural videos is costly.

\section{Dataset}
\label{sec:dataset}

We collected a dataset of abstract scenes to study beliefs of characters. Each scene in our dataset consists of a sequence of $8$ frames showing an everyday situation. One or more people believe something incorrectly about their environment in each scene.
A person may have a false belief for many reasons, including occlusion and misinterpreting intentions.
Although the characters inside the scenes do not know if they are mistaken, we designed the dataset so that third-party viewers can clearly recognize who is mistaken.

Our dataset complements existing abstract scene datasets.
In contrast to the VQA dataset \cite{vqa}, frames in our dataset are grouped into scenes telling stories over several timesteps, and characters in our dataset frequently have mistaken beliefs.


We believe abstract scenes provide a good benchmark for studying visual beliefs. 
We originally tried to collect a dataset of real videos containing people with false beliefs (such as suspense movies), but we encountered significant difficulty scaling up dataset collection. While many real videos contain characters with mistaken beliefs, these beliefs are very complex. This complexity made large-scale annotation expensive. We believe abstract scenes are suitable for understanding visual beliefs today because they allow the field to gradually scale up complexity on this important problem. To recognize mistaken beliefs in real videos, one could always apply domain transfer (e.g.~\cite{ganin2014unsupervised}) to adapt our abstract scenes model to real videos. However, we must first recognize false beliefs in abstract scenes. 

We use our dataset for both learning and evaluation of models for detecting mistaken characters in scenes. 
We show a few examples of our dataset in Figure \ref{fig:grid} and summarize statistics in Figure \ref{fig:data}.
We collected this dataset on Mechanical Turk \cite{sorokin2008utility}. First, we ask workers to illustrate scenes. Then, we ask workers to annotate mistaken characters. In the remainder of this section, we describe how we built this dataset. The appendix contains additional details.


\subsection{Collecting Scenes}

In the illustration step, workers dragged and dropped clipart people and objects into eight frames to tell a coherent story. The interface was a modified version of \cite{vqa}. We told workers that some frame should contain a character who has a mistaken belief about the world. In addition to illustrating these eight frames, workers also wrote a scene-level description and eight frame-level descriptions. These descriptions were used during the annotation step, but were not used to train or evaluate our models.


\subsection{Annotation}

In the annotation step, the goal was to label which characters have mistaken beliefs. We hired workers to review the previously illustrated scenes and write one yes/no question for each frame.
For each frame, workers wrote the true answer to the question and the answer according to each character. We labeled a character as mistaken if their answer was different from the true answer. 

In total, we collected 1,496 scenes, 1,213 of which passed our qualification standards. These scenes were the collective effort of 215 workers. On average, each frame contains 1.71 characters; characters are mistaken in 23.65\% of frames. A pool of 237 workers annotated each scene twice. The labels for whether a character was mistaken were consistent between workers 71.98\% of the time, indicating that in some scenes it was unclear whether a character was mistaken. In this paper, we only consider scenes where characters are clearly mistaken or not.



\subsection{Quality Control}


We used three methods to ensure we collected realistic and diverse scenes.
First, workers completed qualification quizzes before starting the illustration and annotation steps. In the illustration quiz, workers identified good and bad scenes. In annotation quiz, workers filled in characters' answers for a scene with preselected questions. These quizzes forced workers to think about the beliefs of characters. Adding these quizzes significantly increased the quality of our data as compared to a pilot experiment.
Second, the scene background and subset of available people, animals, and objects were randomly selected for each worker, ensuring that workers could not illustrate the same scene twice.
Third, we manually reviewed the first scene illustrated by each worker. If the scene was incoherent or did not contain a mistaken character, we disallowed the worker from illustrating more scenes.

\subsection{What Causes Mistaken Beliefs?} 

 Figure \ref{fig:grid} shows a few scenes from our dataset that highlight different types of mistaken beliefs.
In the first scene, the woman is mistaken because the dog is \textbf{occluded} behind couch, and because she cannot see actions \textbf{outside her field of view}.
In the second scene, the woman falsely accuses the boy of breaking the painting because she cannot observe events when she is \textbf{not present}. 
The girl in the third scene mistakenly assumes the boy can safely get off the teeter totter because of her \textbf{faulty reasoning about physics}.
In the fourth scene, the boy wearing a red shirt \textbf{misinterprets the intentions} of the other boy.
In the last scene, the woman wearing the red shirt lacks the \textbf{common sense} that some mushrooms are poisonous. 
Recognizing mistaken characters requires detecting each of these types of  beliefs.

\section{Belief Tasks}
\label{sec:tasks}

We study two tasks for recognizing mistaken people:

\textbf{Task 1: Who is mistaken?} 
Given a scene and a character, the goal is to predict whether the character is mistaken in any frame. This task has several applications in identifying people who may be confused or unaware of danger.

\textbf{Task 2: When are they mistaken?}
Given a frame, the goal is to predict whether any character is mistaken in this frame. This task has applications in identifying when people might be confused, but it is not possible to know who is confused, such as in a crowd.

\textbf{Joint Task:} We also explore a joint task where we seek to simultaneously recognize who is mistaken as well as localize when they are mistaken in time. 

\section{Method}
\label{sec:method}



We now describe an approach for predicting who is mistaken and when they are mistaken.
Recognizing mistaken characters requires looking beyond a single frame; knowledge of the past or the future can provide important signals for recognizing mistaken beliefs in the present. For example, in the second scene of Figure \ref{fig:grid}, a model must see that the woman was not present when the girl broke the painting to understand why she falsely accused the boy.
Our model for detecting mistaken characters will look at the past, present, and future.
The model must also understand what a person may know and what they might not. To detect a mistaken person, the model should determine that the scene is different from what the person believes.

\subsection{Person-Centric Representation}

\begin{figure}
\centering
\includegraphics[width=0.45\textwidth]{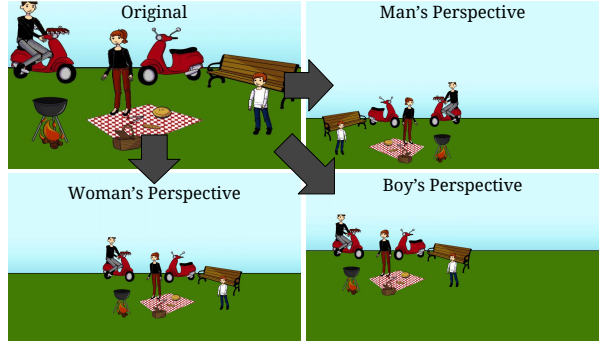}
\vspace{-0.5em}
\caption{\textbf{Person-Centric Representation:} We use a visual representation that focuses on the character of interest.}
\vspace{-1.5em}
\label{fig:egocentric}
\end{figure}

Before predicting whether a character is mistaken, we must tell our model which character to focus on. We use a \textbf{person-centric} representation of the world, where the model takes the perspective of an outside observer focusing on a specific character. For each frame in the scene, we center the frame at the head of the specified character. We also flip the frame so the specified character always faces left. For example, in Figure \ref{fig:egocentric}, the frame in the upper left can be viewed from each of the three characters' perspectives. 
Alternative approaches that remove parts of the frame outside the character's field of view may struggle to reason about what the character cannot see. 



\subsection{Visual Features}
\label{sec:visual-features}

We use a frame-wise approach by extracting visual features for each frame and concatenating them temporally to create a time-series.
We extract visual features from the person-centric images using the AlexNet convolutional network \cite{krizhevsky2012imagenet} trained on ImageNet \cite{deng2009imagenet}. We use activations from POOL5, and further downsample by a factor of two. The resulting feature has size $(256, 12, 21)$. Moreover, although the features we use are trained on natural images (i.e.\ ImageNet), we successfully used them for abstract scenes, possibly because the high rendering quality.


\subsection{Learning}

To learn to predict whether a person is mistaken or not, we can train a regularized convolutional logistic regression model, supervised by annotations from our training set. 
Suppose our image sequences are length $T$ and our features are $D$ dimensional.
Let $\phi(x_i, p_j) \in \mathbb{R}^{T\times D}$ represent the features for sequence $x_i$ for person $p_j$ and $y_{ij} \in \{0, 1\}^T$ be our target category binary, indicating whether person $p_j$ is mistaken in each frame of sequence $x_i$. Our vector of predictions is $\widehat{y}_{i, j} \in \mathbb{R}^T$. We optimize the objective:
\begin{equation}
\begin{aligned}
&\min_w \; \sum_{i, j, t} \left( y_{i,j}^{t} \log(\hat{y}_{i,j}^{t}) + (1-y_{i,j}^{t}) \log(1-\hat{y}_{i,j}^{t}) \right) \\
&\textrm{where} \quad
\hat{y}_{i,j}^{t} = \left( w \ast \phi(x_i, p_j)\right)^{t} + b
\end{aligned}
\end{equation}
The learned weight vector $w \in \mathbb{R}^{K\times D}$ represents the convolutional kernel, where parameter $K$ specifies the temporal width; $b \in \mathbb{R}$ is the learned bias.  For simplicity, we have omitted the L2 penalty on $w$.  The superscript $( \cdot )^{t}$ gives the entry of a vector corresponding to frame $t$ in a scene. We denote convolution as $\ast$, which is performed temporally.  To handle border effects, we pad these features with zeros. The convolutional structure of our model encodes our prior that characters' beliefs are temporally invariant.


\subsection{Who and When}

We tackle two tasks related to beliefs: predict who is mistaken and when they are mistaken. We train a single model that can be used for both tasks. Given a sequence $x_i^t$ centered at time $t$ and a person $p_j$ in the sequence, we train a model to estimate whether person $p_j$ is mistaken at time $t$. To answer the who question, we marginalize the classifier response across time. Likewise, to answer the when question, we marginalize the classifier response across people.

\subsection{Implementation Details}

We extracted image features using Caffe \cite{jia2014caffe} and we used Keras with Theano \cite{bastien2012theano} for learning. To optimize the weights, we used Adam \cite{kingma2014adam}, with a learning rate $10^{-5}$ and a batch size of 32. We set the temporal kernel width $K=7$. We added weight decay with parameter 1, and stopped training after the validation accuracy had stopped increasing for 3 consecutive iterations. Weight decay and downsampling image features helped prevent overfitting.

\section{Experiments}
\label{sec:experiments}

We analyze several models on our dataset of abstract scenes. We evaluate each model on the ``who'' task, the ``when'' task, and the joint ``who + when'' task.

\begin{table}
\centering
\begin{tabular}{l|c|c|c}
& \multicolumn{3}{c}{Task} \\
Method 			& Who+When 		& Who			& When\\
\hline
Chance			& 50 			& 50			& 50\\
Time      		& 62.9 (1.9) 	& 52.4 (1.8)	& 64.3 (2.2) \\
Pose    		& 51.9 (2.1) 	&  50.3 (3.5)	& 54.8 (1.9) \\
Time+Pose  		& 60.6 (2.0) 	& 51.6 (1.2)	& 61.2 (1.9)\\
Facial Expression& 50.1 (1.9)   & 57.4 (5.1)    & 52.9 (2.4) \\
Character ID    & 54.0 (2.1)    & 61.1 (5.4)    & 53.4 (2.4) \\
Present         & 64.5 (2.1)    & 54.1 (6.7)    & 66.1 (2.4) \\
Single Image    & 61.1 (1.7) 	& 59.7 (3.3)	& 62.0 (2.0)\\
Multiple Image  & \textbf{66.6 (1.8)} & \textbf{64.1 (2.8)} & \textbf{67.5 (1.8)} \\
\end{tabular}
\vspace{-0.5em}
\caption{\textbf{Quantitative Evaluation:} We evaluate the accuracy of our model versus various baseline on the who task, the when task, and the joint task. We report classification accuracy; parenthesis show standard deviations.}
\vspace{-1em}
\label{table:main_experiment}
\end{table}

\subsection{Experimental Setup}

We trained each model on the joint task: given a character and a frame, classify if this character is mistaken in this frame. Before training, we balance the dataset by resampling so 50\% of training examples have a mistaken character. We randomly divide the dataset into training/validation/testing splits with sizes 80\%/10\%/10\%. For the experiments in Table \ref{table:main_experiment}, we repeat each experiment 20 times with different splits, and report the mean and standard deviation of the accuracies. For the numbers in Table \ref{table:corrupt}, we only repeat each experiment six times due to cost.

\subsection{Baselines}

We used seven baseline models to study the biases in our dataset, including those shown in Figure~\ref{fig:data}.
We fit a kernelized SVM (RBF kernel) to the three baselines using Time and Pose, use logistic regression for the Single Image model, and use convolutional logistic regression for the  Facial Expression, Character ID, and Present baselines.

\textbf{Time:} This model uses only the time of the frame within the scene, represented as a fraction between 0 and 1.

\textbf{Pose:} This model uses only the pose of the indicated character. Pose includes the $(x, y)$ position of the character, as well as a boolean indicator of whether the character is looking left or right. The $(x, y)$ coordinates are normalized to be in the interval $[0, 1]$.

\textbf{Time + Pose:} This model combines the features from the Time model and the Pose model.

\textbf{Facial Expression:} This model is given only the character's facial expression (encoded as a 1-hot vector).

\textbf{Character ID:} This model is given only the character's identity (encoded as a 1-hot vector).

\textbf{Present}: Each image is replaced by one bit indicating whether the character of interest is present in this frame. To handle border cases, we add another bit to the feature to indicate whether it is padded.

\textbf{Single Image:} This model only looks at the present frame. It is equivalent to our model when $K=1$.


\subsection{Who is mistaken?}

In this experiment, each model is given a scene and a character, and must determine whether the character is mistaken in any frame. The (scene, character) pairs are randomly sampled so 50\%  of pairs contain a mistaken character. If our model only recognized unnatural scenes and ignored the character of interest, it would perform at chance.

We evaluate the model's decision function on each frame in the scene. For the SVM-based baseline models, each prediction is the signed distance from the separating hyperplane; for the models that use logistic regression, each prediction is a value in the interval $(0, 1)$. We take the maximum of these frame-level predictions as the model's scene-level prediction. To obtain a binary decision, we threshold this scene-level prediction (at 0 for the SVM models, and at 0.5 for the logistic regression models).

\begin{table}
\centering
\begin{tabular}{l|c|c|c}
& \multicolumn{3}{c}{Task} \\
Method								& Who+When		& Who 			& When \\
\hline
Chance								& 50			& 50			& 50 \\
Multiple Image						& 66.6 (1.8)	& 64.1 (2.8)	& 67.5 (1.8) \\
\hline
Flipped	                            & 54.5 (1.8) 	& 52.5 (1.7)	& 55.8 (2.4) \\
Centered                			& 62.4 (2.5) 	& 55.6 (3.0)	& 63.0 (2.4) \\
Rewind      						& 57.4 (2.8)	& 61.4 (3.6)	& 57.3 (1.8)
\end{tabular}
\vspace{-0.5em}
\caption{\textbf{Ablation Analysis}: We study the impact of training on altered data and testing on unaltered data. During training, we modify data to flip the character's pose (Flipped), not use the person-centric representation (Centered), and show the frames in reverse order (Rewind). The decrease in accuracy on each task indicates that pose, the person-centric representation, and the arrow of time are important parts of our model.}
\vspace{-1em}
\label{table:corrupt}
\end{table}

\begin{figure*}
\centering
\includegraphics[width=0.9\textwidth]{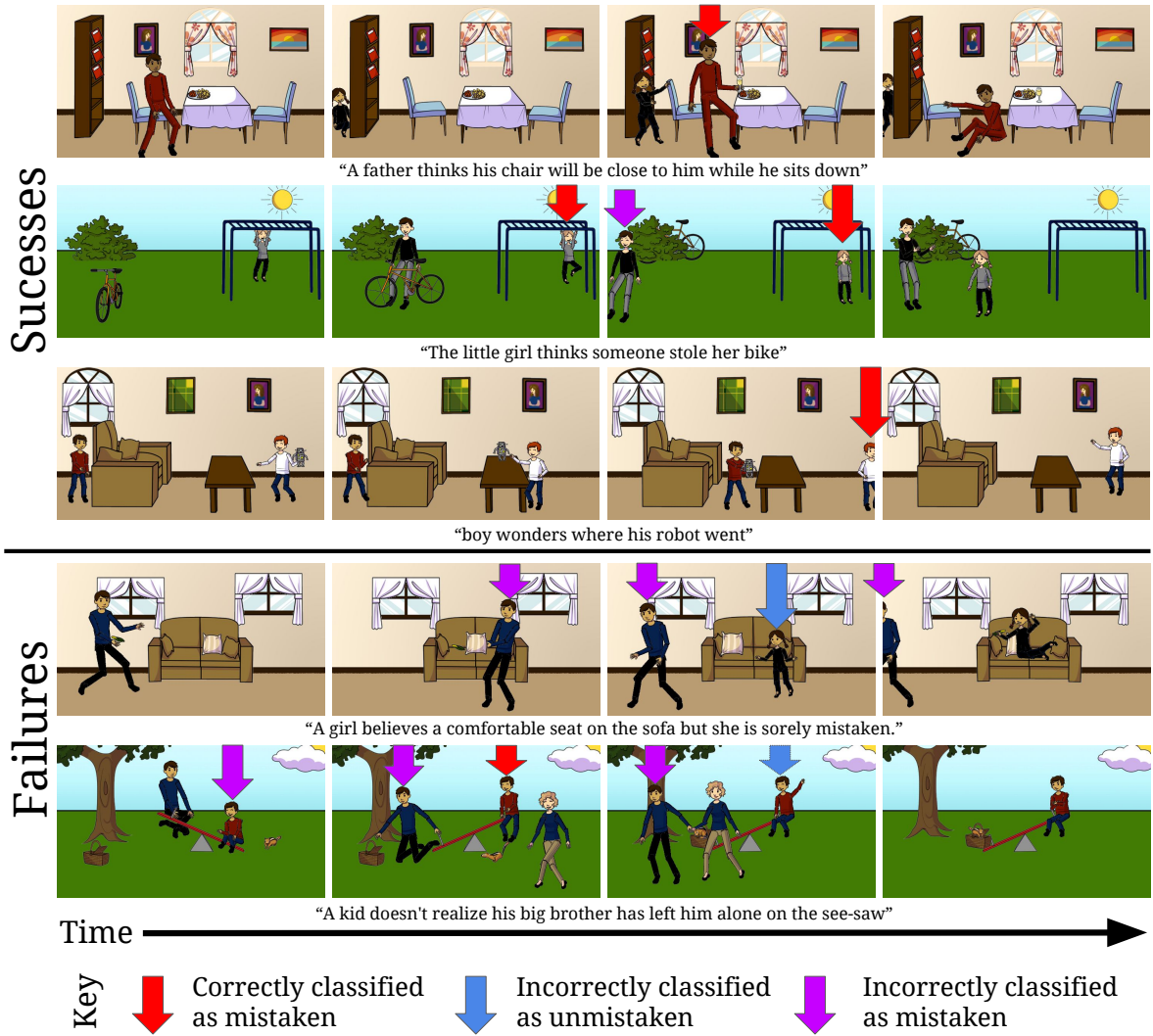}
\vspace{-0.5em}
\caption{\textbf{Example Results:} We show predictions from our model. The first three rows show correct predictions. Our model fails to detect mistaken characters in the last two scenes, which require reasoning about occlusion and physics.}
\vspace{-0.5em}
\label{fig:predictions}
\end{figure*}

The second column of Figure \ref{table:main_experiment} shows that our Multiple Image model achieves a higher accuracy on the ``who'' task than the baselines.
The Facial Expression, Character ID, and Single Image baselines perform better than chance, suggesting that information about the character of interest is important. Our Multiple Image model predicts who is mistaken more accurately than these baselines by also looking at past and future frames.

\subsection{When are they mistaken?}

\label{sec:when-experiment}
In this experiment, each model predicts whether any character in a frame is mistaken.  Frames are randomly sampled so 50\% contain mistaken characters. We evaluate the model's decision function on each character's person-centric representation of the scene. As in the ``who'' experiment, we aggregate predictions across characters by taking the maximum of the model's decision function.

The third column of Table \ref{table:main_experiment} shows that the Time and Present baselines achieve high accuracies, indicating that temporal information is an important for the when task.
The Single Image model performs better than the Pose model, suggesting that the characters' interactions with the scene are important for recognizing mistaken beliefs.
Finally, our Multiple Image model performs better than all baselines. 

\subsection{Joint Task: Who and When?}

In this experiment, the goal is to predict whether a character is mistaken in a given frame.
Frames are randomly sampled so 50\% of (frame, character) pairs contain a mistaken character. As shown in the first column of Table \ref{table:main_experiment}, our model achieves a higher accuracy on the ``who'' task than the baselines. Similar to the ``when'' experiment in Section \ref{sec:when-experiment}, the Time and Present baselines achieve a high accuracies on the joint task. The Pose baseline performs poorly, suggesting that the Time+Pose model likely ignores pose. Although pose is a poor feature for the ``who + when'' task, other features of a single image are important: the Single Image model performs well without knowing the position of the frame in the sequence. The Multiple Image model performs better than all baselines. 

\begin{figure*}
\centering
\includegraphics[width=0.9\textwidth]{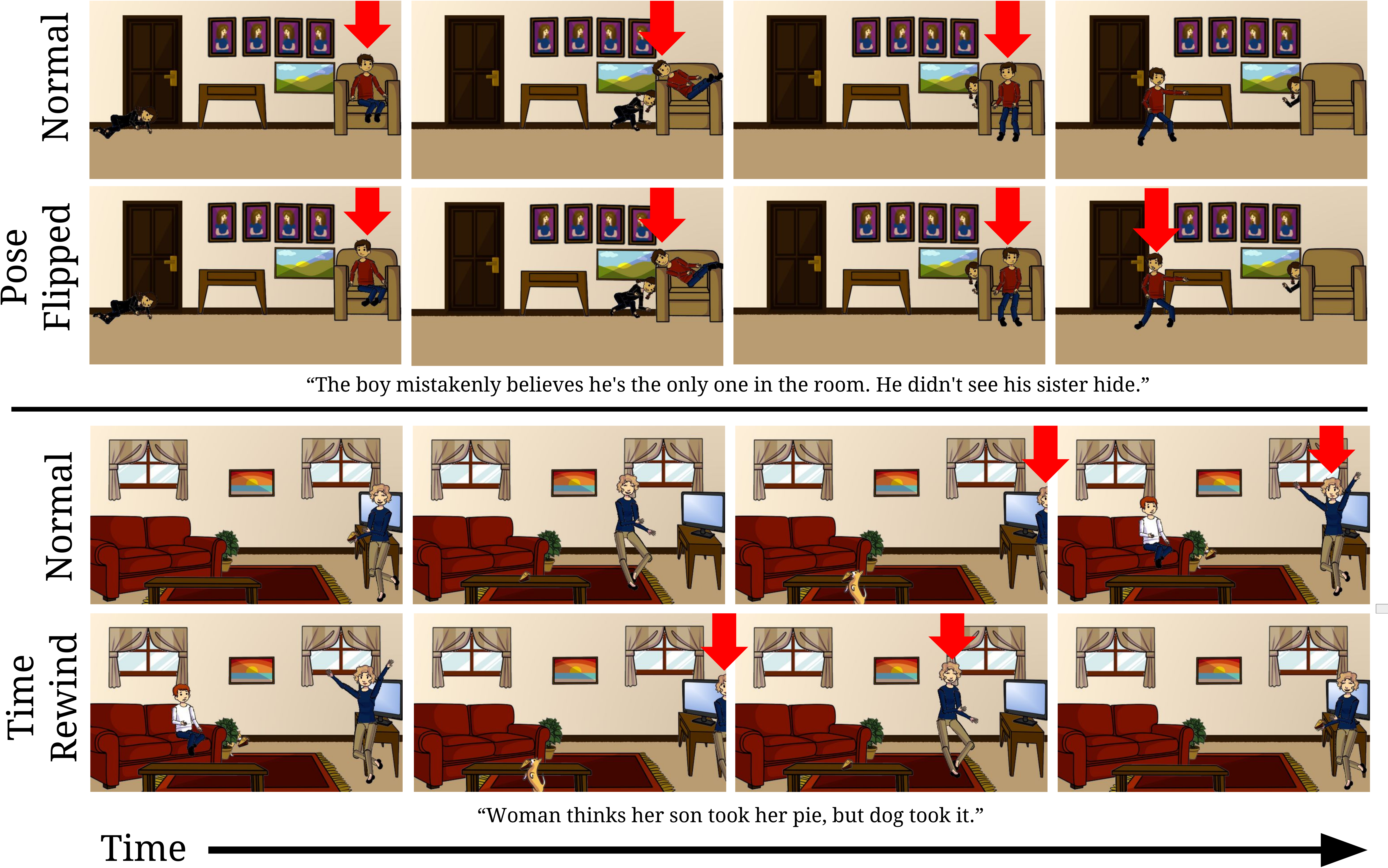}
\caption{\textbf{Predictions from Ablation Experiments:} We visualize our ablation experiments. The first and third rows show a normal scene, and the second and fourth rows show perturbed scenes. \textbf{Row 1:} A normal scene and predictions from our model. \textbf{Row 2:} We flip the boy's pose. In the last frame, the boy no longer sees the girl, so our model predicts he is still mistaken. \textbf{Row 3:} Another normal scene. \textbf{Row 4:} Predictions from the Rewind model make sense for the frames in the fourth row: the woman is mistaken in the second and third frames because she does not see the dog put the pie on the table, and therefore does not know how the pie appeared.}
\vspace{-1em}
\label{fig:ablative}
\end{figure*}

\subsection{Qualitative Results}

Figure \ref{fig:predictions} shows our model's predictions on five scenes. 

\textbf{Row 1:} Our model correctly detects that the man is mistaken in the third frame when the girl is about to pull his chair from beneath him. In this scene, the man is mistaken because he cannot see the girl's actions behind him.

\textbf{Row 2:} Our model correctly predicts that the girl is mistaken in the second and third frames as she can not see the man take her bike. Our model incorrectly predicts that the man is also mistaken in the third frame. Perhaps our model has learned that a character is likely to be mistaken when another character is performing actions behind them.

\textbf{Row 3:} Our model correctly identifies the boy wearing a white shirt as mistaken in the third frame.

\textbf{Row 4:} The man plays a prank on the girl by hiding a piece of corn beneath a pillow. Our model incorrectly predicts that the man is mistaken, likely because he cannot see the actions of the girl behind him. Our model incorrectly predicts that the girl is not mistaken in the third frame, perhaps because the corn is occluded behind the pillow. Our model might think that the corn disappeared when it became occluded. Better models for visual humor could improve our results.

\vspace{1em} 
\textbf{Row 5:} We show another failure case in which a man places a basket on the see-saw, leaving the boy stranded. Here, our model incorrectly predicts that the boy has a misbelief in the first frame, but does not have a misbelief in the third frame. Understanding this situation requires knowledge of basic physics, which our model currently lacks. Advances in physical understanding may improve reasoning about visual beliefs.

\subsection{What has it learned?}

How does our model recognize mistaken characters? In this section, we study some key questions about what our model has learned.

\emph{Does it only detect unusual frames?} Our experiments suggest not. A model for detecting unusual frames would perform well on the when task, but would be unable to do the who task. The Time and Present baselines do well on the when task but poorly on the who task, suggesting that these baselines only detect unusual frames. Our model performs significantly better than chance on the who task, indicating that it does more than detect unusual frames.

\emph{How important is our person-centric representation?} We tested the impact of our person-centric representation by training a \textbf{Centered} version of our Multiple Image model without using the person-centric representation for each character.
As shown in Table~\ref{table:corrupt}, the Centered model performs well on the when task. With no indication of the character of interest, the Centered model performs much worse than our model on the who task, suggesting that our person-centric representation is an important piece of our model.

\emph{Does it do gaze following?} Given that humans use gaze following to reason about the beliefs of others \cite{psych-gaze}, we analyze whether our model started to learn gaze following cues. We trained a \textbf{Flipped} variation on our Multiple Image model that flipped the character's pose during training but not during evaluation.\footnote{We also removed the character of interest from the frame to avoid creating unrealistic images. For example, if we flipped a character sitting on a chair, his limbs would now extend through the back of the chair. We also confirmed that removing the character of interest from our model did not degrade its performance.}

This Flipped model performs worse than our model on the three tasks, as shown in Table \ref{table:corrupt}. This suggests the model is internally learning to use gaze \cite{psych-gaze} without us supervising it to do so. In Figure \ref{fig:ablative}, the top two rows compare predictions made by our original model and the Flipped variation. The predictions made by the Flipped model are consistent with a world where people see from the back of their heads!

\emph{How does it combine information across frames? Does it distinguish between past and future?} Our Multiple Image model outperforms the Single Image baseline, so it must combine information across multiple frames. To investigate how it does this, we ran time backwards during training and forwards during testing. Table~\ref{table:corrupt} shows that this \textbf{Rewind} model performs worse than our model, suggesting that our model treats the past and future differently. 
 In Figure \ref{fig:ablative}, the bottom two rows compare predictions made by our original model and the Rewind variation. The predictions made by the Rewind model are logically consistent if the scene is read backwards (from right to left). This suggests that our model has learned that the arrow of time \cite{pickup2014seeing} is important.

\vspace{-0.5em}
\section{Discussion}
\vspace{-0.5em}

We propose a new computer vision task to recognize when people have mistaken beliefs about their environment. We believe this problem is important because understanding people's beliefs can enable many applications in action prediction, healthcare, and robotics. To spur progress, we introduce a new dataset of abstract scenes to study this problem. We present a model that uses multiple timesteps and a person-centric representation of the scene to recognize mistaken people. Although we only supervise the model with indicators of which characters are mistaken, our ablation experiments suggest that the model learns important cues for this task, such as gaze or the arrow of time.

{\small
\textbf{Acknowledgements:} We thank workers on  Mechanical Turk for their creative scenes. NVidia donated the GPUs used for this research.  This work was supported by a Samsung grant to AT, a Google PhD fellowship to CV, and MIT UROP funding to BE.
}

{\small
\bibliographystyle{ieee}
\bibliography{egbib}

\begin{thebibliography}{10}\itemsep=-1pt

\bibitem{alahisocial}
A.~Alahi, K.~Goel, V.~Ramanathan, A.~Robicquet, L.~Fei-Fei, and S.~Savarese.
\newblock {Social LSTM: Human Trajectory Prediction in Crowded Spaces}.
\newblock In {\em The IEEE Conference on Computer Vision and Pattern
  Recognition (CVPR)}, June 2016.

\bibitem{vqa}
S.~Antol, A.~Agrawal, J.~Lu, M.~Mitchell, D.~Batra, C.~Lawrence~Zitnick, and
  D.~Parikh.
\newblock {VQA: Visual Question Answering}.
\newblock In {\em Proceedings of the IEEE International Conference on Computer
  Vision}, pages 2425--2433, 2015.

\bibitem{bayesian-tom}
C.~L. Baker, R.~R. Saxe, and J.~B. Tenenbaum.
\newblock {Bayesian theory of mind: Modeling joint belief-desire attribution}.
\newblock In {\em Proceedings of the thirty-second annual conference of the
  cognitive science society}, pages 2469--2474, 2011.

\bibitem{bastien2012theano}
F.~Bastien, P.~Lamblin, R.~Pascanu, J.~Bergstra, I.~Goodfellow, A.~Bergeron,
  N.~Bouchard, D.~Warde-Farley, and Y.~Bengio.
\newblock Theano: new features and speed improvements.
\newblock {\em arXiv preprint arXiv:1211.5590}, 2012.

\bibitem{caba2015activitynet}
F.~Caba~Heilbron, V.~Escorcia, B.~Ghanem, and J.~Carlos~Niebles.
\newblock {Activitynet: A large-scale video benchmark for human activity
  understanding}.
\newblock In {\em Proceedings of the IEEE Conference on Computer Vision and
  Pattern Recognition}, pages 961--970, 2015.

\bibitem{castrejon2016learning}
L.~Castrejon, Y.~Aytar, C.~Vondrick, H.~Pirsiavash, and A.~Torralba.
\newblock Learning aligned cross-modal representations from weakly aligned
  data.
\newblock In {\em Computer Vision and Pattern Recognition (CVPR), 2016 IEEE
  Conference on}. IEEE, 2016.

\bibitem{humor}
A.~Chandrasekaran, A.~Kalyan, S.~Antol, M.~Bansal, D.~Batra, C.~L. Zitnick, and
  D.~Parikh.
\newblock {We Are Humor Beings: Understanding and Predicting Visual Humor}.
\newblock {\em arXiv preprint arXiv:1512.04407}, 2015.

\bibitem{chao2015hico}
Y.-W. Chao, Z.~Wang, Y.~He, J.~Wang, and J.~Deng.
\newblock {HICO: A benchmark for recognizing human-object interactions in
  images}.
\newblock In {\em Proceedings of the IEEE International Conference on Computer
  Vision}, pages 1017--1025, 2015.

\bibitem{chen2013neil}
X.~Chen, A.~Shrivastava, and A.~Gupta.
\newblock {Neil: Extracting visual knowledge from web data}.
\newblock In {\em Proceedings of the IEEE International Conference on Computer
  Vision}, pages 1409--1416, 2013.

\bibitem{deng2009imagenet}
J.~Deng, W.~Dong, R.~Socher, L.-J. Li, K.~Li, and L.~Fei-Fei.
\newblock {Imagenet: A large-scale hierarchical image database}.
\newblock In {\em Computer Vision and Pattern Recognition, 2009. CVPR 2009.
  IEEE Conference on}, pages 248--255. IEEE, 2009.

\bibitem{fathi2012learning}
A.~Fathi, Y.~Li, and J.~M. Rehg.
\newblock Learning to recognize daily actions using gaze.
\newblock In {\em European Conference on Computer Vision}, pages 314--327.
  Springer, 2012.

\bibitem{fouhey2014predicting}
D.~F. Fouhey and C.~L. Zitnick.
\newblock Predicting object dynamics in scenes.
\newblock In {\em Proceedings of the IEEE Conference on Computer Vision and
  Pattern Recognition}, pages 2019--2026, 2014.

\bibitem{ganin2014unsupervised}
Y.~Ganin and V.~Lempitsky.
\newblock Unsupervised domain adaptation by backpropagation.
\newblock {\em arXiv preprint arXiv:1409.7495}, 2014.

\bibitem{jia2014caffe}
Y.~Jia, E.~Shelhamer, J.~Donahue, S.~Karayev, J.~Long, R.~Girshick,
  S.~Guadarrama, and T.~Darrell.
\newblock {Caffe: Convolutional architecture for fast feature embedding}.
\newblock In {\em Proceedings of the 22nd ACM international conference on
  Multimedia}, pages 675--678. ACM, 2014.

\bibitem{kingma2014adam}
D.~Kingma and J.~Ba.
\newblock Adam: A method for stochastic optimization.
\newblock {\em arXiv preprint arXiv:1412.6980}, 2014.

\bibitem{kitani2012activity}
K.~M. Kitani, B.~D. Ziebart, J.~A. Bagnell, and M.~Hebert.
\newblock Activity forecasting.
\newblock In {\em European Conference on Computer Vision}, pages 201--214.
  Springer, 2012.

\bibitem{koppula2013learning}
H.~S. Koppula and A.~Saxena.
\newblock {Learning Spatio-Temporal Structure from RGB-D Videos for Human
  Activity Detection and Anticipation}.
\newblock In {\em ICML (3)}, pages 792--800, 2013.

\bibitem{krizhevsky2012imagenet}
A.~Krizhevsky, I.~Sutskever, and G.~E. Hinton.
\newblock Imagenet classification with deep convolutional neural networks.
\newblock In {\em Advances in neural information processing systems}, pages
  1097--1105, 2012.

\bibitem{block-towers}
A.~Lerer, S.~Gross, and R.~Fergus.
\newblock {Learning Physical Intuition of Block Towers by Example}.
\newblock {\em arXiv preprint arXiv:1603.01312}, 2016.

\bibitem{pickup2014seeing}
L.~C. Pickup, Z.~Pan, D.~Wei, Y.~Shih, C.~Zhang, A.~Zisserman, B.~Scholkopf,
  and W.~T. Freeman.
\newblock Seeing the arrow of time.
\newblock In {\em Proceedings of the IEEE Conference on Computer Vision and
  Pattern Recognition}, pages 2035--2042, 2014.

\bibitem{pinto2016curious}
L.~Pinto, D.~Gandhi, Y.~Han, Y.-L. Park, and A.~Gupta.
\newblock {The Curious Robot: Learning Visual Representations via Physical
  Interactions}.
\newblock {\em arXiv preprint arXiv:1604.01360}, 2016.

\bibitem{pirsiavash2012detecting}
H.~Pirsiavash and D.~Ramanan.
\newblock Detecting activities of daily living in first-person camera views.
\newblock In {\em Computer Vision and Pattern Recognition (CVPR), 2012 IEEE
  Conference on}, pages 2847--2854. IEEE, 2012.

\bibitem{prabhakar2010temporal}
K.~Prabhakar, S.~Oh, P.~Wang, G.~D. Abowd, and J.~M. Rehg.
\newblock Temporal causality for the analysis of visual events.
\newblock In {\em Computer Vision and Pattern Recognition (CVPR), 2010 IEEE
  Conference on}, pages 1967--1974. IEEE, 2010.

\bibitem{recasens2015they}
A.~Recasens, A.~Khosla, C.~Vondrick, and A.~Torralba.
\newblock Where are they looking?
\newblock In {\em Advances in Neural Information Processing Systems}, pages
  199--207, 2015.

\bibitem{sadigh2016information}
D.~Sadigh, S.~S. Sastry, S.~A. Seshia, and A.~Dragan.
\newblock Information gathering actions over human internal state.
\newblock In {\em Intelligent Robots and Systems (IROS), 2016 IEEE/RSJ
  International Conference on}, pages 66--73. IEEE, 2016.

\bibitem{scassellati2002theory}
B.~Scassellati.
\newblock Theory of mind for a humanoid robot.
\newblock {\em Autonomous Robots}, 12(1):13--24, 2002.

\bibitem{psych-gaze}
S.~V. Shepherd.
\newblock {Following gaze: gaze-following behavior as a window into social
  cognition}.
\newblock {\em Frontiers in integrative neuroscience}, 4:5, 2010.

\bibitem{sorokin2008utility}
A.~Sorokin and D.~Forsyth.
\newblock Utility data annotation with amazon mechanical turk.
\newblock {\em Urbana}, 51(61):820, 2008.

\bibitem{tzeng2015simultaneous}
E.~Tzeng, J.~Hoffman, T.~Darrell, and K.~Saenko.
\newblock Simultaneous deep transfer across domains and tasks.
\newblock In {\em Proceedings of the IEEE International Conference on Computer
  Vision}, pages 4068--4076, 2015.

\bibitem{abstract-commonsense}
R.~Vedantam, X.~Lin, T.~Batra, C.~Lawrence~Zitnick, and D.~Parikh.
\newblock {Learning common sense through visual abstraction}.
\newblock In {\em Proceedings of the IEEE International Conference on Computer
  Vision}, pages 2542--2550, 2015.

\bibitem{vondrickanticipating}
C.~Vondrick, H.~Pirsiavash, and A.~Torralba.
\newblock Anticipating the future by watching unlabeled video.
\newblock {\em arXiv preprint arXiv:1504.08023}, 2015.

\bibitem{wang2011action}
H.~Wang, A.~Kl{\"a}ser, C.~Schmid, and C.-L. Liu.
\newblock Action recognition by dense trajectories.
\newblock In {\em Computer Vision and Pattern Recognition (CVPR), 2011 IEEE
  Conference on}, pages 3169--3176. IEEE, 2011.

\bibitem{galileo}
J.~Wu, I.~Yildirim, J.~J. Lim, B.~Freeman, and J.~Tenenbaum.
\newblock {Galileo: Perceiving physical object properties by integrating a
  physics engine with deep learning}.
\newblock In {\em Advances in Neural Information Processing Systems}, pages
  127--135, 2015.

\bibitem{xie2013inferring}
D.~Xie, S.~Todorovic, and S.-C. Zhu.
\newblock {Inferring Dark Matter and Dark Energy from Videos}.
\newblock In {\em Proceedings of the IEEE International Conference on Computer
  Vision}, pages 2224--2231, 2013.

\bibitem{yatskarstating}
M.~Yatskar, V.~Ordonez, and A.~Farhadi.
\newblock {Stating the Obvious: Extracting Visual Common Sense Knowledge}.
\newblock In {\em Proceedings of the 2016 Conference of the North American
  Chapter of the Association for Computational Linguistics: Human Language
  Technologies}, pages 193--198, San Diego, California, June 2016. Association
  for Computational Linguistics.

\bibitem{zhang2015yin}
P.~Zhang, Y.~Goyal, D.~Summers-Stay, D.~Batra, and D.~Parikh.
\newblock {Yin and Yang: Balancing and answering binary visual questions}.
\newblock {\em arXiv preprint arXiv:1511.05099}, 2015.

\bibitem{tom-hri}
Y.~Zhao, S.~Holtzen, T.~Gao, and S.-C. Zhu.
\newblock {Represent and Infer Human Theory of Mind for Human-Robot
  Interaction}.
\newblock In {\em 2015 AAAI Fall Symposium Series}, 2015.

\bibitem{zitnick2013bringing}
C.~L. Zitnick and D.~Parikh.
\newblock Bringing semantics into focus using visual abstraction.
\newblock In {\em Proceedings of the IEEE Conference on Computer Vision and
  Pattern Recognition}, pages 3009--3016, 2013.

\end{thebibliography}
}

\clearpage
\appendix

\section*{Appendix}

In this appendix, we provide more details on how workers illustrated and annotated our dataset. We also animate our scenes.

\section{Illustrating the Dataset}
\label{sec:illustration}

\begin{figure}[!h]
    \begin{subfigure}[b]{0.45\textwidth}
    \includegraphics[width=\textwidth]{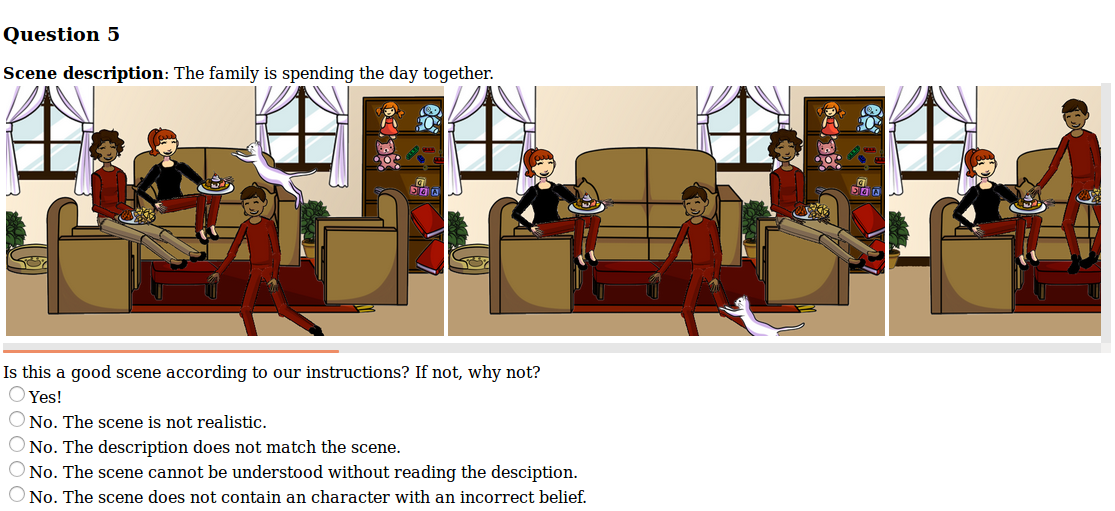}
    \end{subfigure}
    \begin{subfigure}[b]{0.45\textwidth}
    \includegraphics[width=\textwidth]{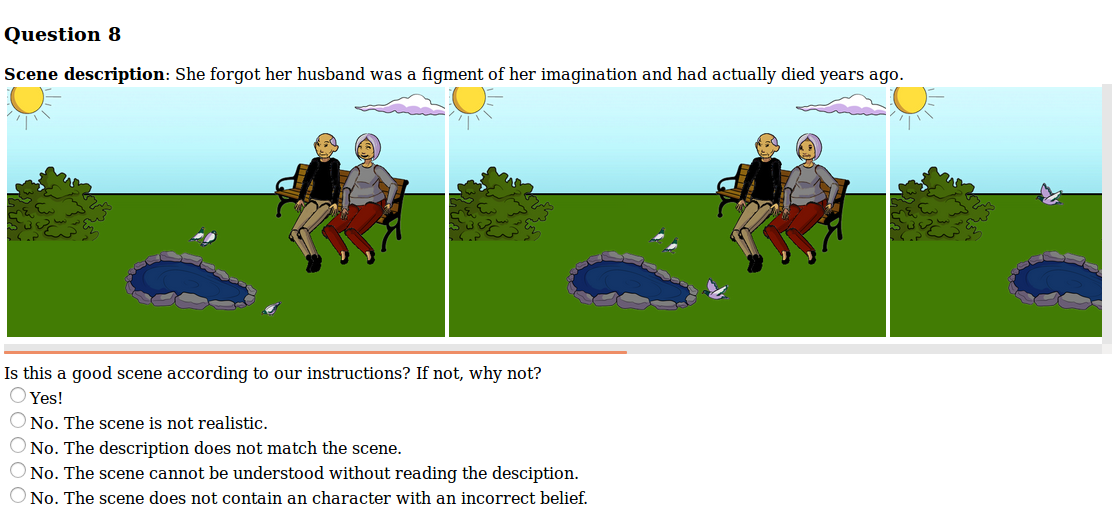}
    \end{subfigure}
    
    \caption{\textbf{Illustration quiz:} Two scenes from the illustration quiz. Workers scrolled left/right to see the 8 frames in each scene. \label{fig:illustration-quiz}}
    \vspace{-1em}
\end{figure}

\subsection{Illustration Quiz}
The first time workers logged in, they were presented with a quality control quiz. In this quiz, workers were shown a number of scenes, and were asked ``Is this a good scene according to our instructions? If not, why not?'' Workers chose one of the following options:
\vspace{-0.5em}
\begin{itemize}
    \item Yes! \vspace{-.8em}
    \item No. The scene is not realistic. \vspace{-.8em}
    \item No. The description does not match the scene. \vspace{-.8em}
    \item No. The scene cannot be understood without reading the description. \vspace{-.8em}
    \item No. The scene does not contain a character with an incorrect belief.
\end{itemize}

Figure \ref{fig:illustration-quiz} shows some of the scenes from our quiz. Note that the workers could scroll left/right to see all eight frames in the scene. Workers could begin illustrating their own scenes only after correctly completing this quiz.

The scenes shown in the illustration quiz were chosen to highlight common mistakes we saw in a small pilot experiment we ran prior to collecting the main dataset. We found that adding the quiz significantly improved the quality of scenes in the main dataset as compared to the pilot experiment.

\subsection{Illustration Interface}

\begin{figure}
\centering
\includegraphics[width=0.45\textwidth]{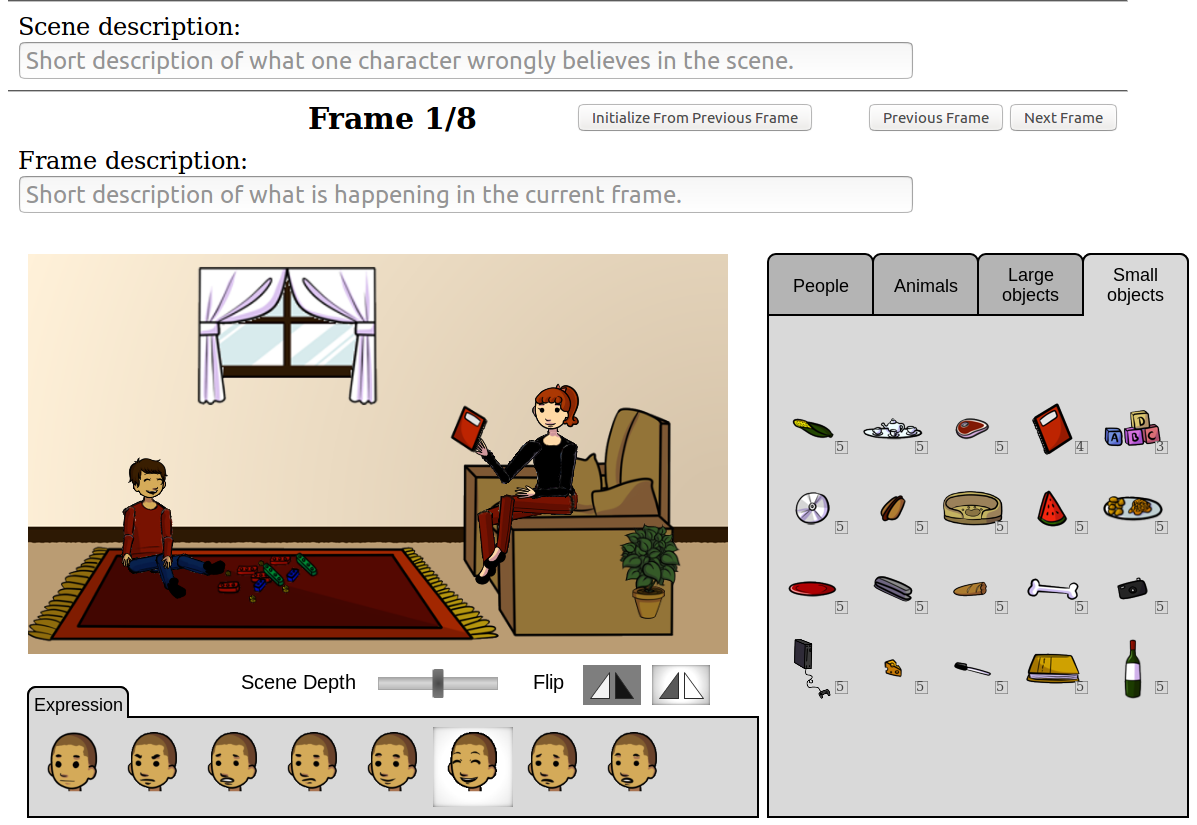}
\caption{\textbf{Illustration interface:} This is the tool workers used to illustrate scenes. The people, animals, and objects available in the right pane were chosen randomly to diversify our dataset.}
\vspace{-1em}
\label{fig:illustration-interface}
\end{figure}

Figure \ref{fig:illustration-interface} shows the illustration interface. There are four tabs to the right of the scene for choosing people, animals, large objects, and small objects to add to the scene. After illustrating a scene, workers also provided a scene-level description and eight frame-level descriptions. These descriptions were used to help workers annotate our dataset, but were not used to train our model.

\section{Annotating the Dataset}
\label{sec:annotation}


\begin{figure*}[h!]
    \begin{subfigure}[b]{0.45\textwidth}
    \includegraphics[width=\textwidth]{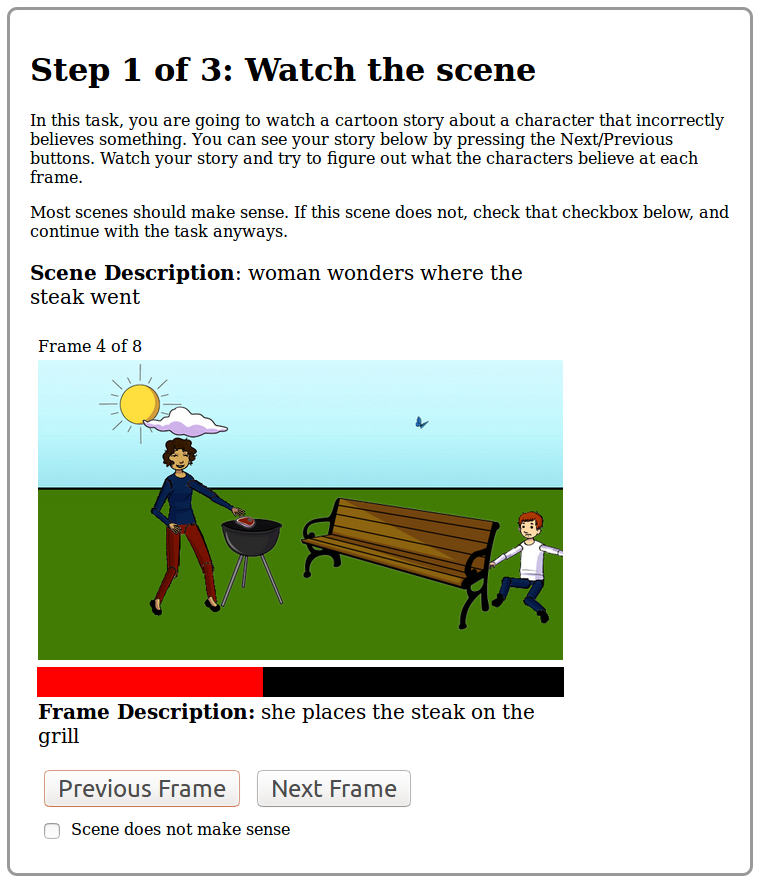}
    \end{subfigure}
    \hfill
    \begin{subfigure}[b]{0.45\textwidth}
    \includegraphics[width=\textwidth]{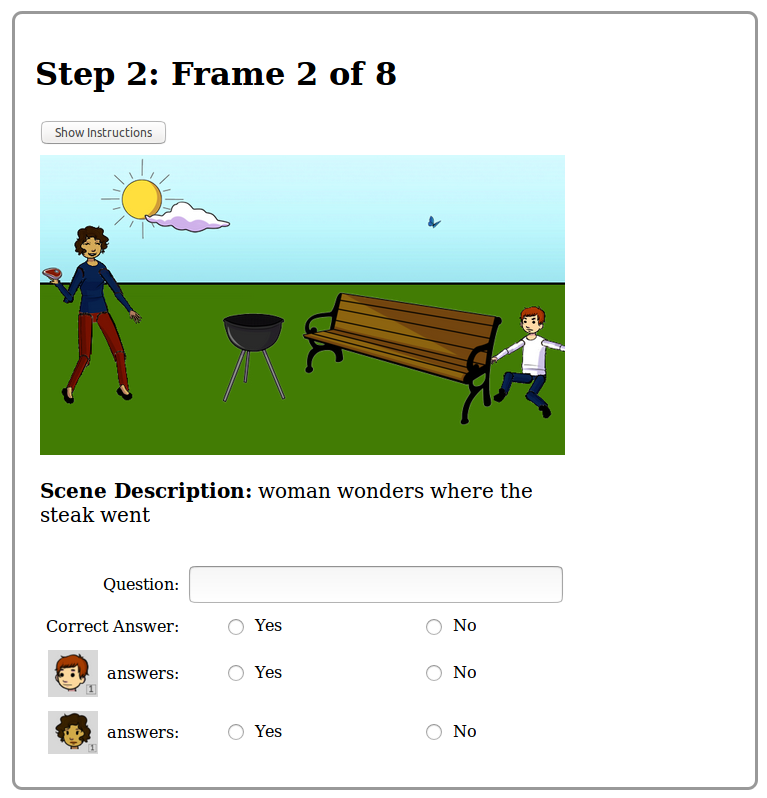}
    \end{subfigure}    
    \caption{\textbf{Annotation interface:} In the first part of the annotation step (\textbf{left}), workers studied the scene. In the second part (\textbf{right}), workers wrote questions and answers about each frame. These questions and answers were used to determine which characters were mistaken. The third step (not shown) allowed workers to submit feedback.
    \vspace{-1em}
    \label{fig:annotation-interface}}
\end{figure*}

\begin{figure}[h!]
    \centering
    \includegraphics[width=.45\textwidth]{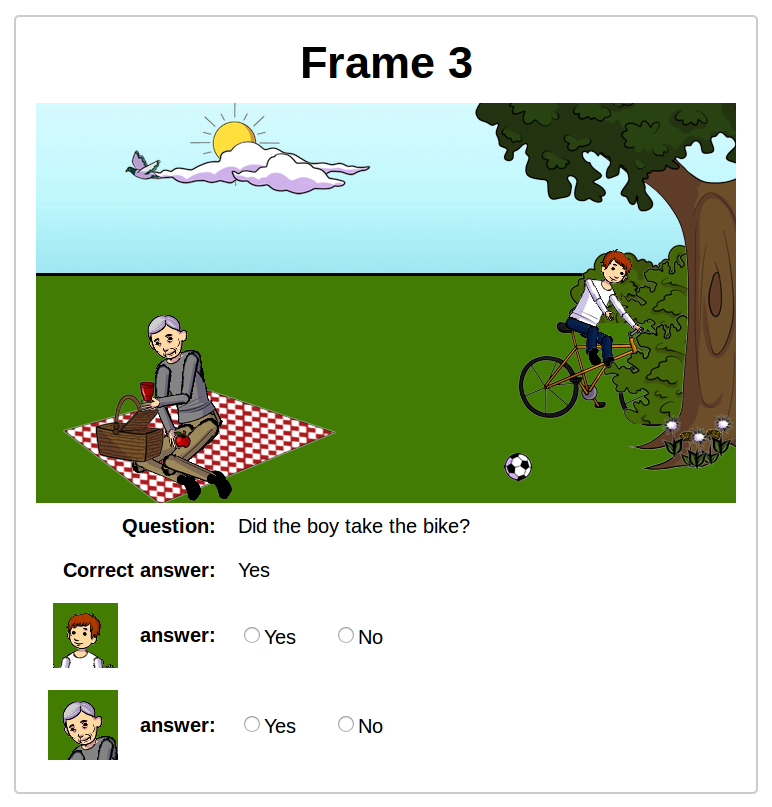}
    \caption{\textbf{Annotation quiz:} One example from the annotation quiz. Workers saw the entire 8-frame scene when answering questions about this frame. \label{fig:annotation-quiz}}
    \vspace{-1em}
\end{figure}

\subsection{Annotation Quiz}
Before workers could start annotating scenes, they completed a short quiz. In this quiz, we showed workers a couple scenes and accompanying questions. The workers were asked how each character in each scene would answer the question. Figure \ref{fig:annotation-quiz} shows two frames from two scenes in the quiz. Workers saw all 8 frames for each scene. In the frame on the left, the boy would answer ``yes'' because the boy knows he (the boy) did take the bike; the man would answer ``no'' because he thinks the boy did not take the bike.









\subsection{Annotation Interface}
After completing the annotation quiz, workers annotated scenes from our dataset. First, workers studied the scene, as shown in Figure \ref{fig:annotation-interface} (left). Second, workers wrote a question that some character would answer incorrectly in some frame, as shown in Figure \ref{fig:annotation-interface} (right). Workers also predicted how each character would answer the question in each frame. Note that these questions and answers were only used to identify mistaken characters. Their text was not used to train our model. 

We showed the annotators both the scene-level description and the frame-level descriptions. These helped annotators understand the problem we were studying. Importantly, the scenes were illustrated so it is possible to understand the scenes without reading these descriptions.

\section{Animation}
\label{sec:animation}

To provide another way of understanding our dataset, we animated the scenes. Because we have access to the generative parameters for each scene, it is easy to interpolate between frames. Note that the interpolated frames were not used to train our model. Rather, these videos highlight how access to the generative parameters are a unique strength of our dataset. These videos can be seen on the project webpage: \mbox{\small{\url{http://people.csail.mit.edu/bce/mistaken/}}}.

\end{document}